\documentclass[10pt,a4paper,onecolumn]{article}
\usepackage{epsfig,subfigure, float, amsmath, xtab}
\usepackage{cite}

\begin{document}

\title{Robot Companions: Technology for Humans}

\author{Serge Kernbach\footnote{In: Jeremy Pitt (Edt.) This Pervasive Day: The Potential and Perils of Pervasive Computing, Imperial College Press, 2012.}\\
\small Institute of Parallel and Distributed Systems, University of Stuttgart, \\
\small Universit\"atstr. 38, 70569 Stuttgart, Germany, \\
\small email: {\it serge.kernbach@ipvs.uni-stuttgart.de}}

\date{}
\maketitle

\section{Introduction}\label{sec1.1}

Creation of devices and mechanisms which help people has a long history. Their inventors always targeted practical goals such as irrigation, harvesting, devices for construction sites, measurement, and, last but not least, military tasks for different mechanical and later mechatronic systems. Development of such assisting mechanisms counts back to Greek engineering, came through Middle Ages and led finally in XIX and XX centuries to autonomous devices, which we call today "\emph{Robots}"~\cite{Craig05}. \index{robot}

\begin{figure}[htp]
\centering
\subfigure[\label{fig:Karakuri}]{\epsfig{file=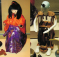,width=.55\textwidth}}~
\subfigure[\label{fig:ASIMO}]{\epsfig{file=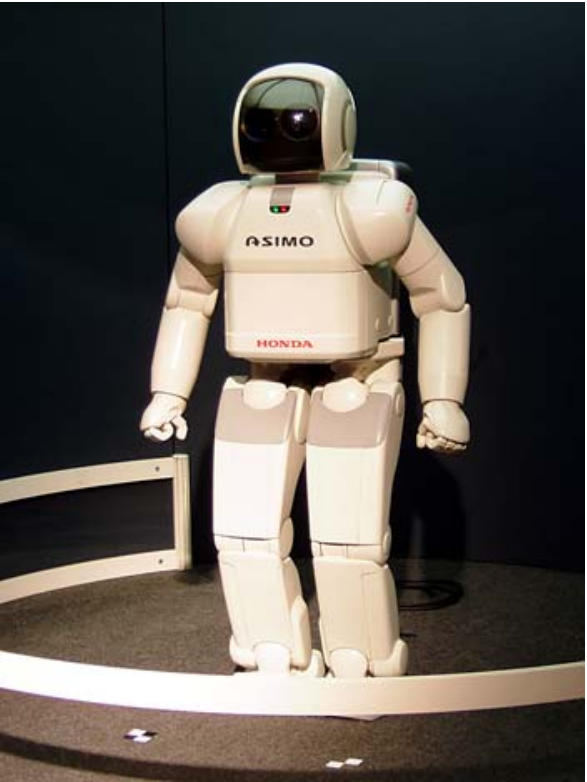,width=.4\textwidth}}
\subfigure[\label{fig:jasmine}]{\epsfig{file=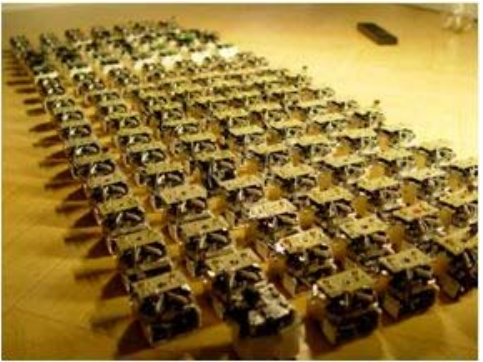,width=.512\textwidth}}~
\subfigure[\label{fig:bacteria}]{\epsfig{file=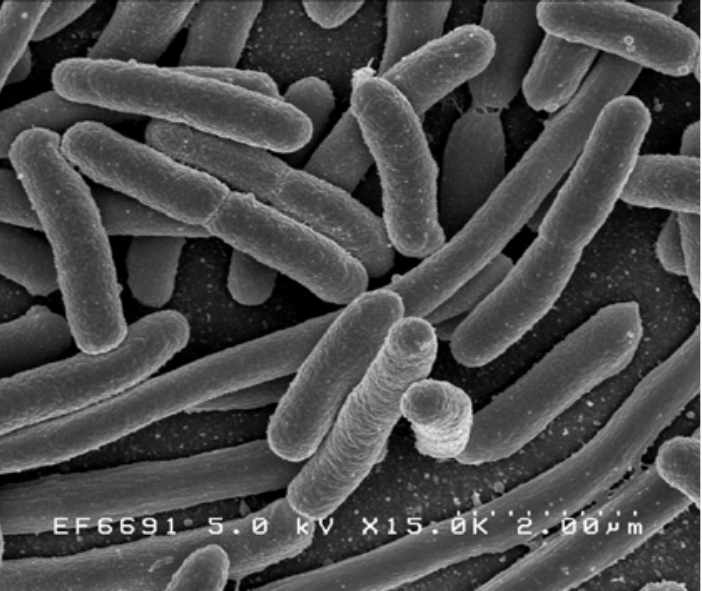,width=.452\textwidth}}
\caption{Different examples of robots. \textbf{(a)} Karakuri ningyo, japan mechanical automata from 18 and 19 century (image source: Wikipedia Commons); \textbf{(b)} Humanoid robot ASIMO created by Honda (image source: Wikipedia Commons); \textbf{(c)} Collective robotics, swarm robots Jasmine; \textbf{(d)} Bacterial robotics: Scaning electron micrograph of Escherichia coli, grown in culture and adhered to a cover slip (images source: Wikipedia Commons).\label{fig:robots}}
\end{figure}

\emph{Robot} embodies three essential principles: mechanical strength due to utilization of independent energy source, programmable actuation through perception and information processing, and autonomy of behavior. All these principles can be implemented in different ways, for instance Japan's karakuri ningyo, mechanical automata from XVIII -- XIX centuries, used mechanical energy and specific wood gears~\cite{Law97}, see Fig.~\ref{fig:Karakuri}. Industrialization replaced water and mechanical tension energy by steam, chemical and electrical energy; this resulted in a new generation of mechanisms. The term of \emph{"robot"} appeared first in 1920 in a science-fiction novel~\cite{Zunt07} and reflected a public discussion of that time related to development of such assisting devices.

Modern robotics has almost hundred years of development from the first mechatronic devices of XX century. It includes now such branches as industrial, service, entertainment and educational \index {robotics} robotics; differentiated into humanoid (see Fig~\ref{fig:ASIMO}), field and collective systems (see Fig~\ref{fig:jasmine}); covers water, surface, air and space domains; it is represented in nano-, micro-, mini- and macro- worlds from $10^{-7}$ m. to $10^{2}$ m. sizes~\cite{Siciliano08}, \cite{KernbachHCR11}. Robotic technology advanced into micro-mechatronics, genetic engineering, bottom-up chemistry and bacterial (see Fig~\ref{fig:bacteria}) areas. Currently, robotics is integrated not only into a human society but also exists as mixed robot-animal systems and even raises the question of independent robot cultures~\cite{WinfieldGriffiths10}.

Human-robot interaction is one of critical issues of modern robotics. It comes back to the origin as an assisting device for humans and requires re-considering the old dilemma: do we need "friend, assistant or butler"~\cite{Dautenhahn05}? does a robot companion aim only at elderly people~\cite{aisb-05}? can a robot be a mate? Robot companions for humans require re-thinking human aspects in human-robot interactions and estimating a boundary of our acceptance of technology. Are we ready to accept a synthetic life or we need only an assisting device? This chapter is intended to overview a technological transition \emph{autonomous device} $\rightarrow$ \emph{humanoid assistant} $\rightarrow$ \emph{social robots}, which robotics undergoes now and to focus on \emph{social} and \emph{self-driven} developmental aspects of robot technologies.

This chapter is structured in the following way: Section~\ref{sec:STA-robot-technology} briefly introduces different state-of-the-art robot technologies in hardware and software areas. Section~\ref{sec:future-robot} is devoted to future technologies, which are potentially applicable in robotics in next few years and Section~\ref{sec:collective-robotics} overviews a large field of collective systems. Section~\ref{sec:social} discusses social aspects of robot companions and driving forces of modern robotic development. Finally, Section~\ref{sec:conclusion} returns back to the topic of \emph{This Perfect Day}, mentioned in the introduction to this book, polemises self-driven trends in robotics and concludes this chapter.

\section{State-of-the-art Robot Technologies}
\label{sec:STA-robot-technology}

\begin{figure}[htp]
\centering
\subfigure{\epsfig{file=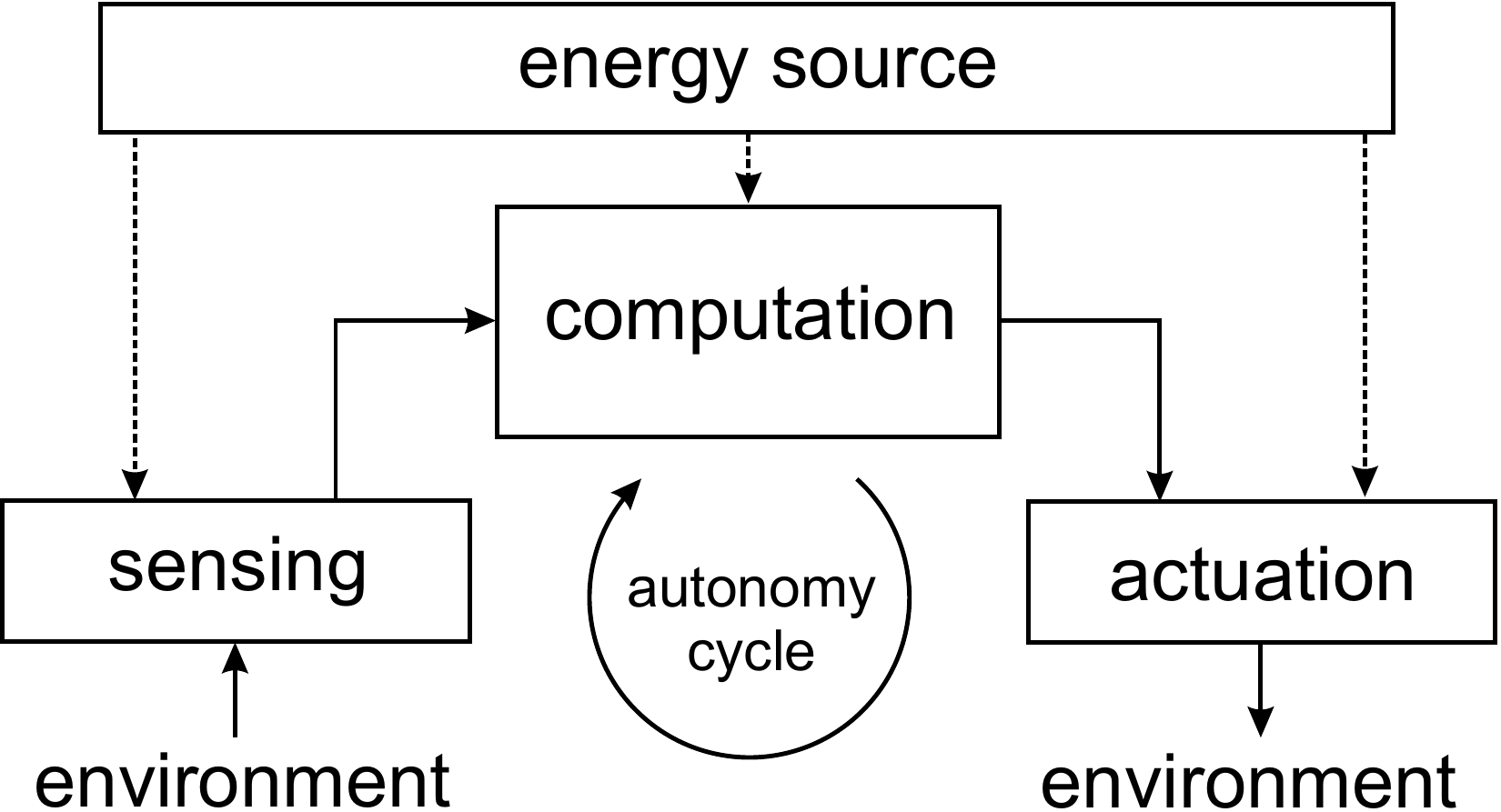,width=.6\textwidth}}~
\caption{General dependence between sensing, actuation, energy and computation parts. Cyclical execution of sensing, computational and actuating tasks is denoted as the autonomy cycle. \label{fig:scheme}}
\end{figure}

To make overview of different robot technologies, we may look first for a definition of ``robot'', which is commonly defined as:
\begin{itemize}
\item
\textbf{The free on-line dictionary:} ``\emph{a mechanical device that sometimes resembles a human and is capable of performing a variety of often complex human tasks either on command or by being programmed in advance.''}
 \item
\textbf{Word History:} \emph{``robot'' comes from Czech ``robota'', meaning ``servitude" or ``forced labor", derived from rab, "slave". The Slavic root of ``robota" is ``orb-," from the Indo-European root *orbh-, referring to separation from one's group or passing out of one sphere of ownership into another.}
\item
\textbf{The Robotics Institute of America:} \emph{``a robot is a reprogrammable multifunctional manipulator designed to move materials, parts, tools, or specialized devices through variable programmed motions, for the performance of a variety of tasks".}
\end{itemize}
or, to generalize, a system capable of:
\begin{enumerate}
\item  on-board sensing;
\item  on or off-board autonomous or semi-autonomous data processing;
\item  on-board energy supply or on-board energy transformation;
\item  actuation and/or interactions with its environment.
\end{enumerate}
The dependence between (1)-(4) is shortly sketched in Fig.~\ref{fig:scheme}. Cyclical execution of sensing, computational and actuating tasks is denoted as the \emph{autonomy cycle}. To some extent, robotics can be defined as an integration science over sensing, actuation, energy and computation areas.
However, it not only follows technological developments from other domains, due to autonomy cycle and independent energy source, robots include \index{behavioral robotics} \emph{autonomous and behavioral aspects}. Both are two important features and lead to an appearance of \emph{cognitive functionality}. Further, a combination of these three properties and involvement of different learning approaches contribute to \index{artificial evolution} \emph{evolutionary and developmental features}. Thus, four interacting hardware elements (sensing, actuation, energy and computation) enable five "soft features" (autonomy, behavior, adaption, cognition, and development): all of them determine modern mechatronic robotics.

\begin{figure}[htp]
\centering
\subfigure{\epsfig{file=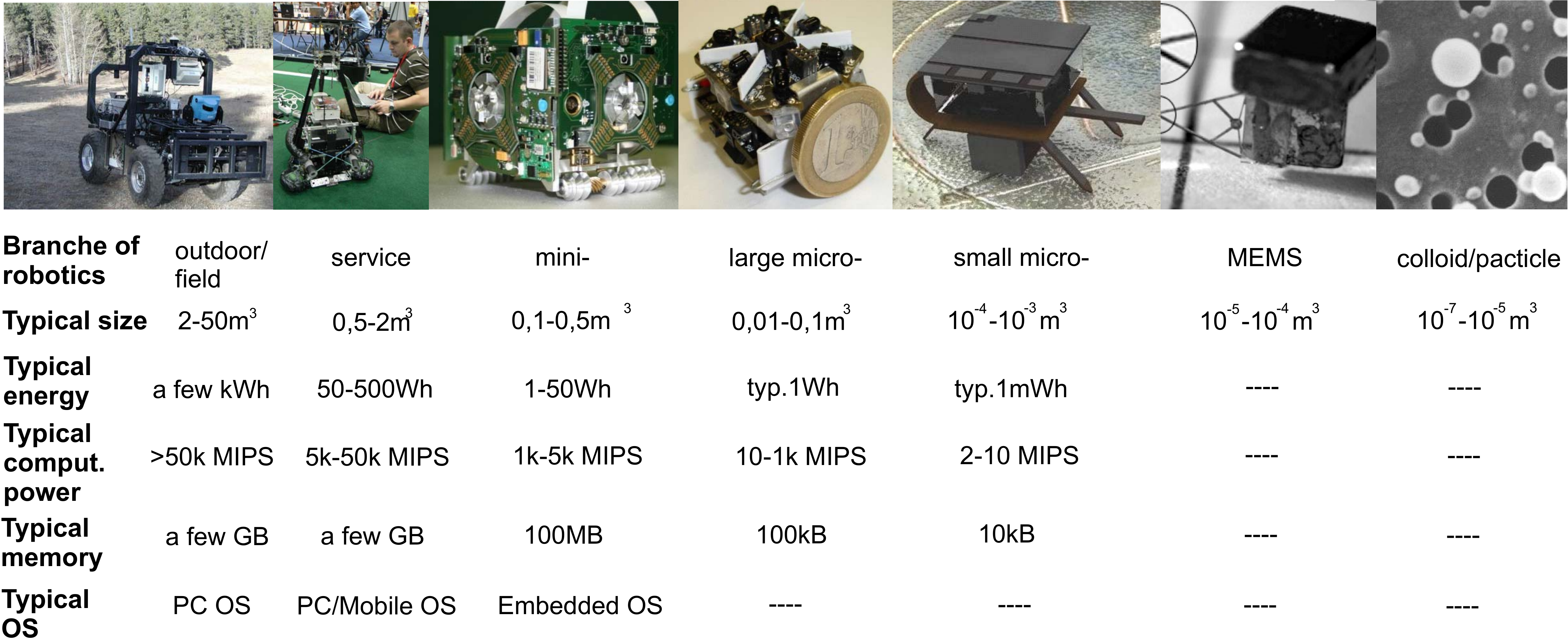,width=1.\textwidth}}
\caption{Seven robot classes based on the dependency between available on-board energy, computational power and size of robots. Typical values/examples of size, available on-board energy, computational power, memory and operating systems (OS) are shown for different classes of mobile robot platforms. \label{fig:robotsOver}}
\end{figure}

There are several robot taxonomies~\cite{Siciliano08}; based on such criteria as size, available onboard energy and on-board computational power, we can roughly define seven classes of robot platforms, shown in Fig.~\ref{fig:robotsOver}.
In following, we make a short overview of robot technologies for these classes of robots.

\textbf{Energy.} \index{energy harvesting} On-board energy supply is one of the most challenging problems in mobile robotics. Generally, a dependency between the size of a mobile system and available energy defines robot classes, shown in Fig.~\ref{fig:robotsOver}. With the reduction of system's size, the energy consumption is reduced, however the available place for energy is also reduces. For the large field platforms, combustion engines are used, whereas smaller platforms use mostly rechargeable electrical energy sources, such as lithium-polymer accumulators. Due to autonomous recharging capabilities, autonomous work in almost all platforms varies between a few hours and a few days. There are several projects which explore alternative energy sources, for example Microbian~\cite{Aaron10:MFC} or Hydrogen~\cite{Patil04} Fuel Cells technologies as well as utilization of energy harvesting~\cite{Priya08}.

\textbf{Actuation.} Actuation is the most energy consuming operation, typically 70\%-80\% of on-board energy is spent for actuation. Therefore, actuating capabilities are closely related to the available energy for each class of robot systems from Fig.~\ref{fig:robotsOver}. Generally, there are only a few types of actuators used in robotics: magnetic/electrostatic actuators, DC, AC and piezzo-motors, pneumo- and hydraulic- actuators. Their capabilities also vary from "very strong" in field devices to "very weak" in micro-systems. Actuation and sensing are closely related to each other: when sensing enables a robot to perceive the environment, due to actuation a robot can modify the environment. Both are used for direct and indirect communication, for example trophallaxis \index{trophallaxis} and \index{stigmergy} stigmergy.

\textbf{Information processing.} Available energy and size define finally computational capabilities of robots: from several PCs on-boards with thousands MIPS up to tiny embedded systems with only a few MIPS in 8bit microcontrollers. Due to development of micro-electronics and smart mobile devices, on-board information processing essentially advanced in last years in all classes of robots. Primarily this technology is related to microcontollers, memory, peripheral control and buses. The level of current integration in Programmable Systems on Chip provides all needed functionalities in one chip with flexible reconfiguration of components.

\textbf{Sensing.} Sensing technologies in robotics follow general trends from industrial sensor development. They use common optical, sonar or electromagnetic approaches with transmissive and reflective techniques~\cite{KornienkoS05d}. Application of vision-based methods for navigation and localization is also well-established. The sensing range varies between "very large" in field robotics up to "very small" in micro-systems. Normally, sensor data require preliminary processing, such as noise reduction or linearization, and final data processing, where data are matched with the model to underhand the meaning the current sensor information. These capabilities in turn depend on the available computational power.

\textbf{Autonomy and Behavior.} As mentioned above, through a cyclic execution of sensing, computational and actuating tasks, a robot exhibits a phenomenon, which we call autonomous behavior. Autonomy \index{autonomy} is very basic property, we say that \textit{autonomy in robotics is a capability to act independently as a self-determining closed system}. Degree of robot autonomy caused a large discussion in robotic, multi-agent and human-robot interaction communities~\cite{Goodrich07}. The concept of autonomy can separately be applied to sensing, computation and actuation, so that two principally different viewpoints on autonomy can be encountered. Firstly, a robot  keeps its autonomy, in other words, it performs sensing, data processing and actuating autonomously. We say, this is a \emph{strong autonomy}. Secondly, sensing, actuation and computation can be shared among different robots. One or several processes can be performed on the level of the whole system or even on a host-computer. Robots still retain some degree of autonomy, however they depend on other agents in decision making. We denote this autonomy as a soft or \emph{weak autonomy}.

Autonomy enables a robot to behave as an \emph{independent unit}. Since robots operate in environments, they interact with other robots and also with environments. These interactions introduce behavioral aspects into robotics. Depending on the level of central or decentralised coordination~\cite{Kornienko_S03A} and the degree of autonomy, the \index{robot interactions} interactions between robots can demonstrate several kinetic relationships~\cite{Levi99}, geometrical and spatio-temporal dependencies~\cite{CIM06}, energetic balance with the environment, and coevolution with each other and with the  environment~\cite{Futuyma83}. For example, some behavioral aspects of swarm systems are similar to the behavior of gas molecules and multi-particle systems and these similarities have stimulated several macroscopic probabilistic approaches to modeling~\cite{martinoli1999b}. Robotic systems are often used to model and to investigate different biological systems~\cite{Schmickl08}.

\textbf{Adaptation, Cognition and Development.} These three concepts are high-level issues in robotics. Adaptability \index{adaptability} is closely related to environmental changes, the ability of a system to react to these changes, and the capability of the designer to forecast the reaction of the environment to the system's response. Therefore, adaptability is defined in term of the triple-relationship: \emph{environmental changes}$\rightarrow$ \emph{system response} $\rightarrow$ \emph{environmental reaction}. In general, adaptability is the ability of a system to achieve desired environmental reactions in accordance with a priori defined criteria by changing its structure, functionality or behavior~\cite{kernbach09adaptive}. Adaptive technical systems are expected to have some degrees of freedom, so they may adapt to their environment. In this context, adaptivity is closely related to three issues: developmental plasticity, capability to detect changes and, finally, mechanisms allowing reaction to changes by utilizing plasticity. Since adaptive systems are approached from several independent directions, the understanding of these underlying mechanisms differs from community to community.

\emph{Cognition} \index{cognition} is a general term, which implies a perception, processing of information, building and changing of knowledge and their expressions in behavior~\cite{Indiveri_etal09}. When firstly the notion of cognition was developed in psychology, a development of autonomous behavioral systems leaded to a transfer of these principles to robotics~\cite{AndyClark99}. Since the cognition unifies other notions such as autonomy, behavior or adaptation, currently robotics undergoes a "re-thinking process" as embodied cognitive science.

\emph{Development} \index{development} is the most highest notion and has its origin in biological systems as an ontogenetic development of an organism, i.e. from one cell to a multi-cellular adult system~\cite{Spencer08}. The development in this context is related to epigenetic systems, which "explain how phenotypic characteristics arise during development through a complex series of interactions between genetic program and environment"~\cite{Brauth91}. Plasticity of development~\cite{Kernbach08_2} is related to cause-effect sequences by which information is read out in genotype in the presence of environmental stimuli. Artificial developmental systems, in particular developmental (epigenetic) robotics~\cite{LungarellaMPS03}, are new emerging fields across several research areas -- neuroscience, developmental psychology, biological disciplines such as embryogenetics, evolutionary biology or ecology, and engineering sciences such as mechatronics, on-chip-reconfigurable systems or cognitive robotics~\cite{Asada2009-CDR}. We can say that the development is the bounded or unbounded process of functional, structural and regulatory changes undertaken by the system itself, related to its specific understanding of itself (expressed by so-called self-concept). Normally, the development or self-development is initiated by differences between the self-concept and endogenous or environmental factors and may be unlimited in time and complexity.

Five mentioned "soft components" of modern robotics have different expression in seven classes of robot platforms. First of all, the high-level functionality requires a lot of computational resources, which are not available on small platforms. Secondly, the operating environment has an essential impact on needs of higher cognitive capabilities: requirements in unstructured outdoor environments are much harder than those in domestic and laboratory conditions. In this way, we observe currently two different trends: working with minimalistic capabilities, for example minimally cognitive systems~\cite{DaleH10}, and a full scale approach with extended sensing and actuation.

\section{Bio-, Chemo- and Neuro- Hybrid Robotics}
\label{sec:future-robot}

Classical mechatronic devices, overviewed in the previous section, represent now only a part of robotics. We face now new developments from synthetic biology, molecular~\cite{Balzani03} bacterial~\cite{Sylvain09}, colloidal~\cite{Hunter89}, bio-hybrid~\cite{Novellino07} and cultured neural~\cite{Reger00} systems, artificial chemistry and self-replication~\cite{Hutton09}. It seems that robotics should be defined in more broad way, without emphasizing the mechatronic point of view. Overviewing the state of art in hybrid systems, we can currently differentiate four mainstreams in development of hybrid systems: bio-techno artefacts, neuro-hybrid, chemo-hybrid and bio-hybrid \index{hybrid robotics} systems.

\begin{figure}[htp]
\centering
\subfigure[\label{fig:hybro1}]{\epsfig{file=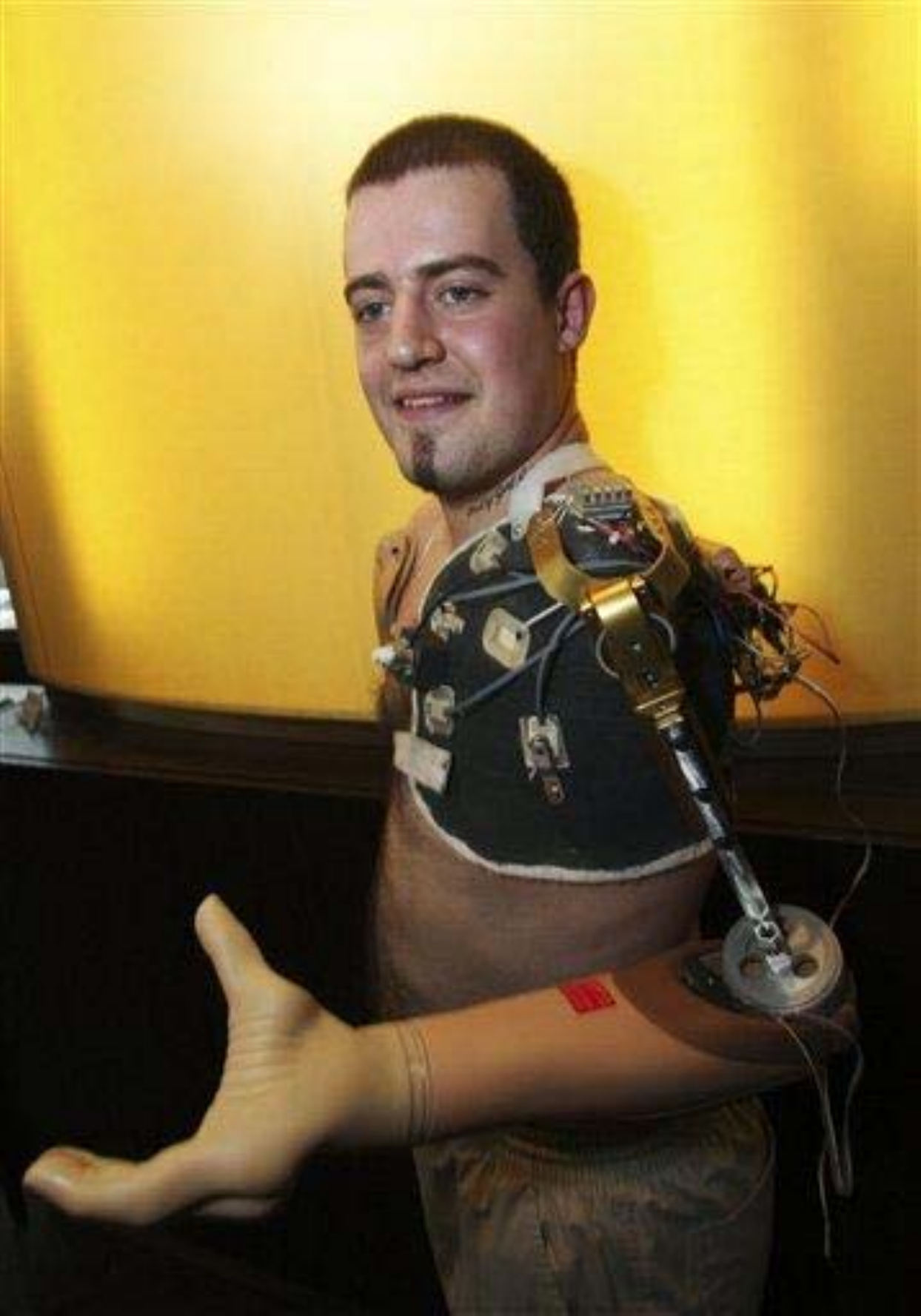,width=.35\textwidth}}~
\subfigure[\label{fig:hybro2}]{\epsfig{file=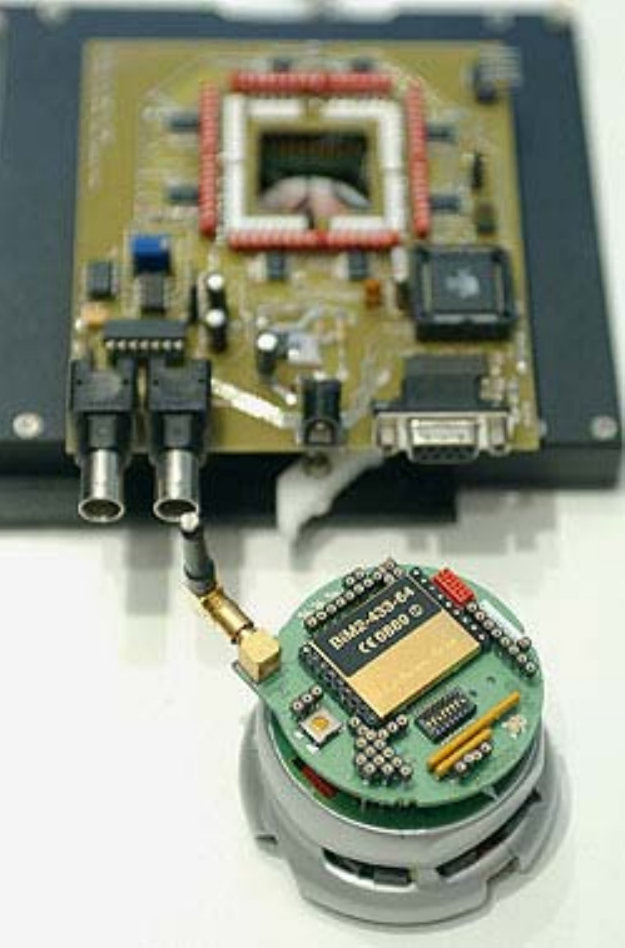,width=.33\textwidth}}~
\subfigure[\label{fig:hybro3}]{\epsfig{file=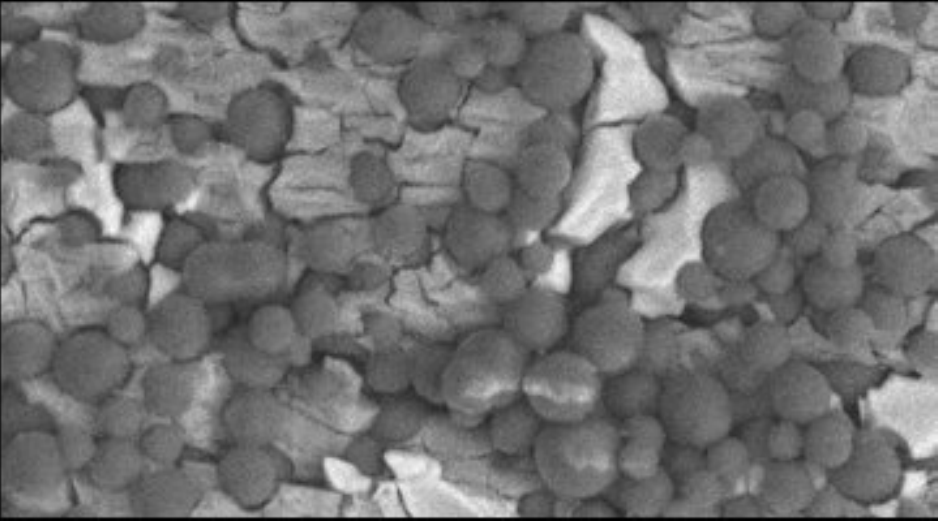,angle=90,width=.28\textwidth}}
\caption{\textbf{(a)} Christian Kandlbauer lost both arms in an electrical accident in 2005 but was able to live a largely normal life thanks to a mind-controlled robotic prosthetic left arm and a normal prosthesis in place of his right arm (with permission, AP Photo/Ronald Zak); \textbf{(b)} Hydrot robot~\cite{DeMarse01}, a robot controlled through biological neurons (images source: Wikipedia Commons); \textbf{(c)} Prototypes of chemical robots "chobots" (images source: BBC News; credit Frantisek Stepanek);
\label{fig:hybrid}}
\end{figure}

Hybrid robotics addresses several important questions, one of them is an attempt to interact with biological systems by means of \index{technological artefacts} technological artefacts. Examples can be given by robotic \index{robotic prostheses} prostheses, see e.g. Fig~\ref{fig:hybro1}, controlling mixed societies of robot and insects~\cite{caprari05}, autonomous management of the grazing cattle over large areas~\cite{SchwagerJFR08}, or by social communication between robots and chickens~\cite{Gribovskiy09}. A similar approach is related to the integration of different robot technologies into human societies, for example the management of urban hygiene based on a network of autonomous and cooperating robots~\cite{Mazzolai08}.

Interactions between neuroscience and robotics represent another interesting research area in hybrid systems. This is primarily related to a combination of cultured (living) neurons and mobile robots~\cite{Novellino07} to investigate the dynamical and adaptive properties of neural systems~\cite{Reger00}. This work is also related to the understanding of how information is encoded~\cite{Cozzi06}, and processed, within a living neural network. For instance, the Hybrot project~\cite{DeMarse01}, see Fig.~\ref{fig:hybro2} or the NeuroBit project~\cite{MartinoiaNeuroBit04} addressed the problem of the control of autonomous robots by living neurons. This hybrid technology can be used for neuro-robotic interfaces, different applications of \emph{in vitro} neural networks~\cite{Miranda09}, or for bidirectional interaction between the brain and the external environment.

Another approach in hybrid systems is inspired by artificial chemistry~\cite{Dittrich01}, self-replicating systems~\cite{Hutton09}, using bio-chemical mechanisms for, for example, cognition~\cite{Dale10}, or the well-known quasispecies~\cite{EigenM_1971}. In several works, this approach is denoted as swarm chemistry~\cite{Sayama09}, see Fig.~\ref{fig:hybro3}. Researchers hope that such chemistry-based systems will give answers to questions related to developmental models~\cite{Astor00}, chemical computation~\cite{Berry92}, self-assembly, self-replication, and simple chemistry-based ecologies~\cite{Breyer97} of prebiotic life.

Bio-hybrid approach is represented by synthetic biology and the integration of real bio-chemical and microbiological systems into technological developments; for example using bacterial cellular mechanisms~\cite{Wood99} as sensors, the development of bacterial bio-hybrid materials, the molecular synthesis of polymers~\cite{Pasparakis10} and biofuels~\cite{Alper09}, genome engineering~\cite{Carr09}, and more general fields and challenges of synthetic biology~\cite{Alterovitz09}. Here, bio-hybrid and pure bio-synthetic systems overlap in several applications; we expect that both areas will be developed in parallel.

Bio-, chemo- and neuro- hybrid \index{neurorobotics} systems raise new technological challenges to robotics: the integration of bio-chemistry, micro-biology, synthetic biology, and robotics will be a vital challenge in coming years and promises several radical breakthroughs regarding the adaptive and developmental properties of artificial systems. Currently there are only a few real applications for them (such as robotic prostheses), however we expect that in future this field will be one of the dominant areas of robotics as an assisting technology for humans.

\section{Collective Robotics}
\label{sec:collective-robotics}

\begin{figure}[htp]
\centering
\subfigure[\label{fig:collective3}]{\epsfig{file=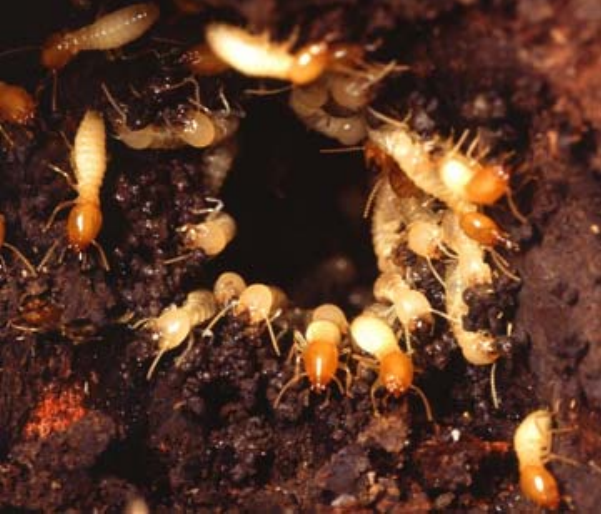,width=.435\textwidth}}~
\subfigure[\label{fig:collective4}]{\epsfig{file=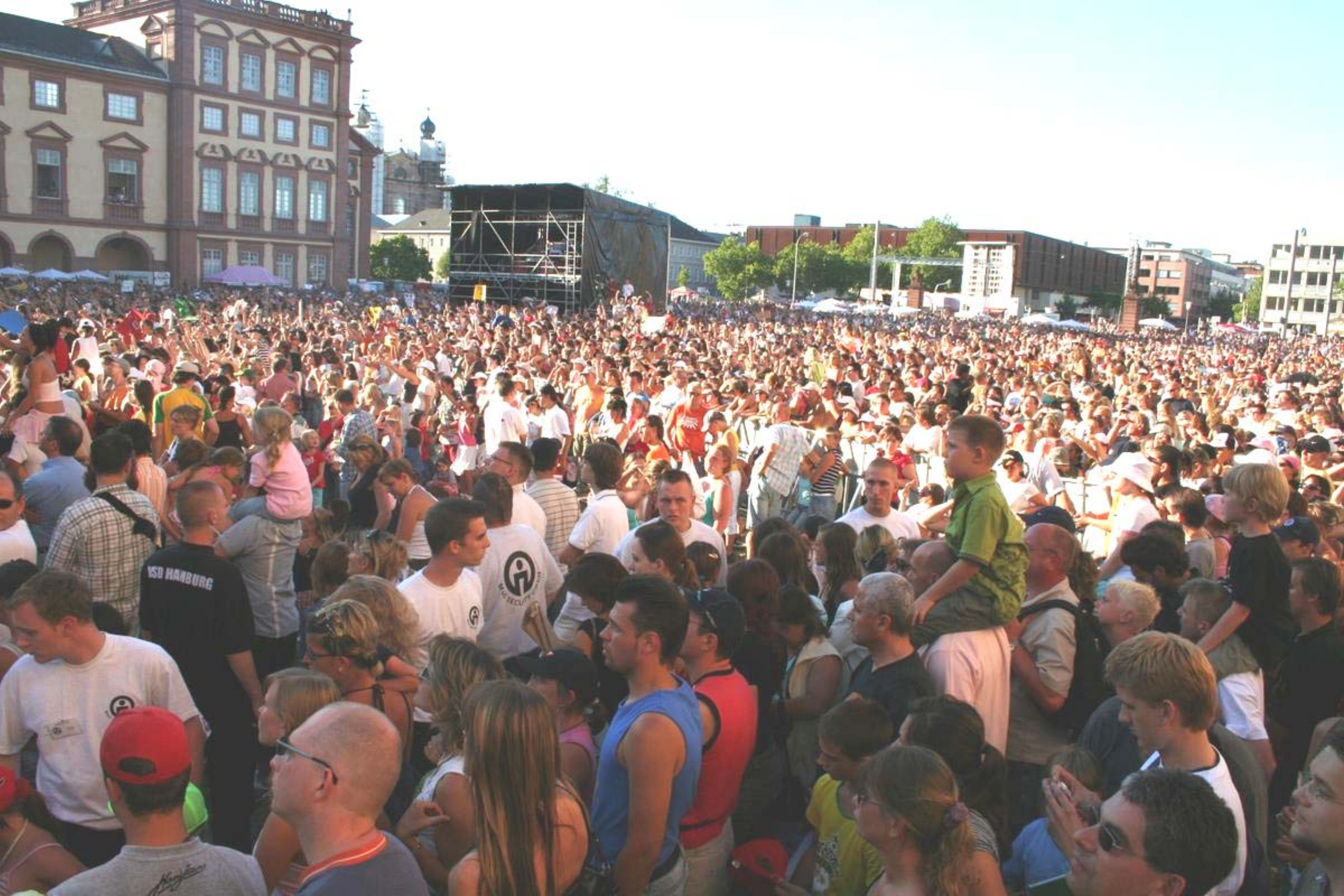,width=.56\textwidth}}
\subfigure[\label{fig:collective1}]{\epsfig{file=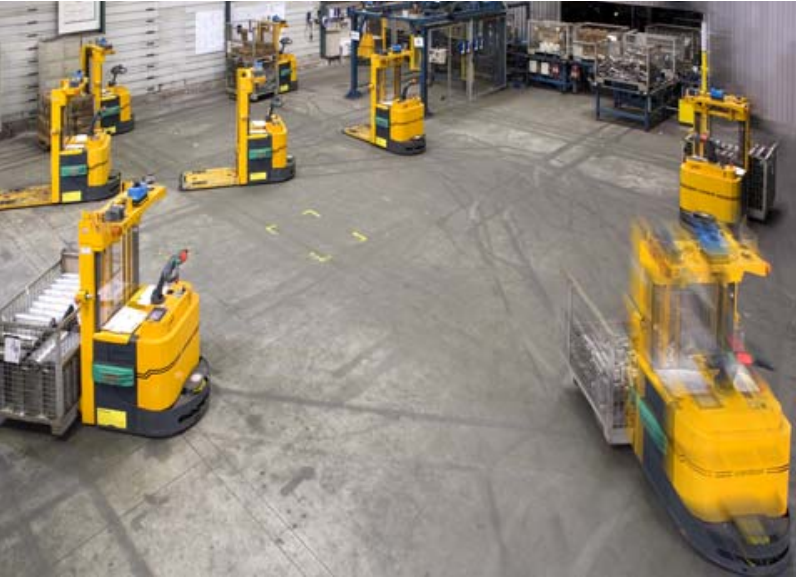,width=.46\textwidth}}~
\subfigure[\label{fig:collective2}]{\epsfig{file=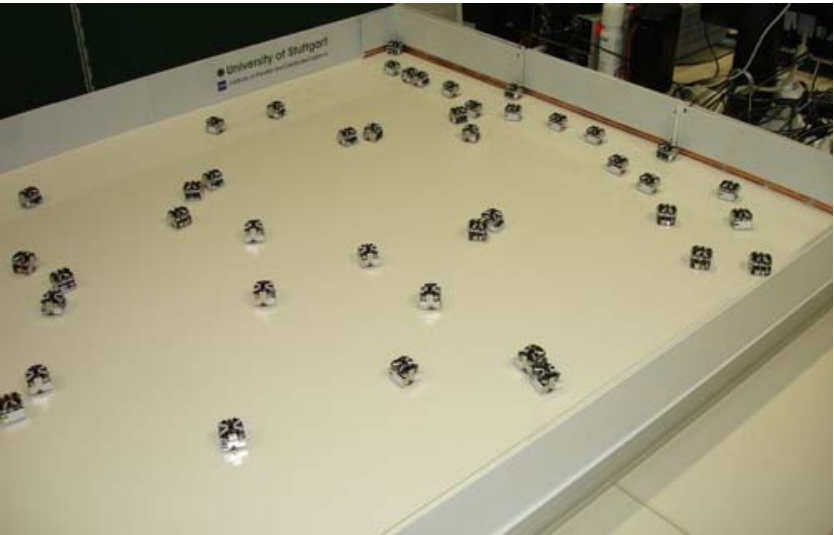,width=.52\textwidth}}
\caption{Examples of collective systems. \textbf{(a)} Termites (images source: Wikipedia Commons);  \textbf{(b)} Human crowd (images source: Wikipedia Commons); \textbf{(c)} Automatic guided vehicles, courtesy of psb intralogistics GmbH; \textbf{(d)} Swarm robots;
\label{fig:collective}}
\end{figure}

There are two reasons, why \index{collective robotics} collective robotics should be mentioned as a separate area of research. Firsts of all, collective robotics in many aspects overlaps with a larger domain of collective systems, which include many biological, bio-chemical or sociological examples, see Figs.~\ref{fig:collective3} and~\ref{fig:collective4}. Problems appeared in such populations can be modeled and analyzed with robots~\cite{Kernbach08online}. Secondly, one of the most common features of collective systems is the idea that by ``working together'', such systems achieve results which are not attainable by individuals alone~\cite{Kornienko_S04b}. Generally, collective systems consist of many interacting \emph{individuals}, such as molecules, insects, animals, robots, software agents, or even humans. Through their \emph{interactions}, the individuals of a collective system are able to achieve behaviors, functionality, structures and other properties that are not achievable by these individuals alone. Such synergetic affects are of essential interest in many other disciplines~\cite{Haken77}, which do not touch robotics directly.

Collective robotics represents a relatively new area of artificial systems. The earliest references go back to the late 80s and early 90s~\cite{fukuda89}. Collective robotic systems appear when sensing, data processing and actuation are distributed among many different robots. We know ``the system is more than the sum of its parts''~\cite{Aristothel89}; this \emph{``more"} -- often referred as the common organizational principle -- is the most important issue when designing collective autonomous systems. The common organizational principle can be understood as a common goal, common strategy or something that makes the system focused on one. This defines how the sensing and computational processes are distributed among different ``pieces of hardware and software'': ``\emph{Given some task specified by a designer, a multiple robot system displays cooperative behavior if, due to some underlying mechanism for example the ``mechanism of cooperation'', there is an increase in the total utility of the system.}''~\cite{cao97}.

Thus, collective robotics is an intentionally designed system of interacting autonomous or semi-autonomous robots, which distribute or share sensing, computation, energy supply or actuation. These robots can be of technological, bio-synthetic, or any other origin, and with a common organizational principle such as a common goal, intention or strategy underlying the whole system. In most cases, a collective robot system consists of many independent autonomous individual robots, each of which is capable of sensing, computation and actuation. Sometimes, these robots can be autonomous in actuation, however they may share computational and  sensing resources, as do those in the cooperative  KUKA's ``RoboTeam" robots~\cite{Vasilash05}. In some cases, robots are not fully autonomous, but are still distributed in actuation, such as automatic guided vehicles (AGV)~\cite{Watanabe01}, see Fig.~\ref{fig:collective1}. Recently, collections of a large number of relatively simple robots have come to represent so-called swarm robotics, see Fig.~\ref{fig:collective2}, which is historically situated in observation of the natural world, in particular social insects.

The concept of collective systems provides several essential advantages~\cite{Kornienko_OS01}. First, systems consisting of many independent autonomous elements are very reliable. Second, collective systems have many degrees of freedom and are much more flexible than centralized ones. Such flexibility can be used for adaptation~\cite{Kornienko_S05e}, or for developmental processes. Third, due to the decentralization of regulative and functional strictures, collective systems are scalable in a wide range of structural, diversity and dynamic conditions~\cite{Constantinescu04}.

\section{Social Aspects of Robotics and Robot Companions}
\label{sec:social}

Technologies, shortly mentioned in the previous sections, define many facets of modern robotics. However, following the original idea of a robot as an assisting technology to human, we need to investigate another side of this problem -- the need of human for assistance. The bridge between current capabilities of robots and this need for assistance is addressed within the area of human-robot interactions and design of sociable robots~\cite{Breazeal02}. Considering the Fig.~\ref{fig:robotsOver}, we can say that different classes of robots are capable of performing different tasks, and first of all, we need to clarify a potential usage of these classes. In the table below we briefly summarize several of these classes, their main characteristics and possible applications.

Based on this table, we can identify several key drivers of robotics as the technology for human. First of all, it needs to keep separated (a) potentially interesting developments, which may have an essential role in future, (b) various academic research programs, and finally (c) industrial applications.
\vspace{2mm}

\xentrystretch{-0.1}
\begin{xtabular}{p{1,7cm}@{\extracolsep{3mm}}p{4,8cm}@{\extracolsep{3mm}}p{4,6cm}}\hline
Class of platforms & Several Characteristics & Potential Usage\\\hline
Field robotics
&\small general class of large outdoor robots of $2-50m^3$ and $>kWh$ onboard energy; operation in unstructured and dynamic environments.
&\small construction, forestry, agriculture, mining, sea and underwater, search and rescue, military, and space. \\
Outdoor robotics
& \small different, primarily mobile systems from field- and service- class platforms and intelligent vehicles, targeting specific challenges of outdoor environments.
& \small different outdoor applications, autonomous transportation of persons and goods.
\\
Service robotics
&\small general class of indoor and outdoor robots, intended to use in different service areas. It varies typically between 0,5-2$m^3$ size and 50Wh-500Wh on-board energy.
&\small large application domain related to human companions, e.g. guide in museum, hospitals, different assisting tasks.
\\
Domestic robotics
& \small part of service robotics, which is related to "@home applications"; dynamic home-like indoor environments with fixed structure.
&\small assistance at home, different sorts of cleaning, intelligent refrigerators, ironing and similar.
\\
Medical and Health Care robotics
& \small different systems, such as manipulators, preprogrammed or teleoperated robots, operating in very specific medical environments.
& \small surgery, orthopedics, endoscopic
surgery, microsurgery, rehabilitation tasks, physical therapy and training.
\\
Mini- robotics
& \small general class of small indoor robots. It varies typically between 0,1-0,5$m^3$ size and 1Wh-50Wh on-board energy.
& \small domestic and educational applications, different service tasks in small areas.
\\
Micro- robotics
& \small general class of very small robots. It varies typically between 0,0001 - 0,1 $m^3$ size and 0,001Wh - 1Wh on-board power.
& \small several medical applications, education, micro- and nano- manipulation, fluidic systems, factories on the table.
\\
Colloidal
and Particle
systems
& \small new class of "robots" with programmable properties. It consists of several different systems of biological, chemical or material science origin.
& \small programmable bio-chemical and hybrid systems, drug design and discovery, applications in molecular manufacturing.
\\\hline
\end{xtabular}
\vspace{2mm}

This progress from (a) to (c) is challenging and obviously not all results can be transferred, however it clearly indicated an enhancement of initial "toy-scenarios" to real products. Below we show four such application areas.

\textbf{Industrial Needs.} This one of the main current drives in many different areas of automatization and robotics and is primarily related to handling, processing, assembling or transportation. To some small extend, toy industry also stimulates development of robot technologies. New \index{industrial applications} industrial applications in such areas as drug discovery and drug design involve robotics. Potentially interesting for industrial applications are micro- and nano-technological systems, dealing micro-actuation, micro-cleaning, structurization of material and similar tasks. Here we can also mention automotive industry and attempts of creating autonomous intelligent vehicle as well as diffident military applications of this technology. Space robotics for autonomous space missions or space manufacturing, can also be located in the sector.

\textbf{Future computation, surveillance, S\&R, sensor networks.} These are today's applications, where small- and middle-size groups of robots perform tasks of inspecting or monitoring technological processes, checking availability of specific/dangerous substances, tracking objects/persons. Depending on the number of robots and their degree of autonomy, these applications vary between mobile computing devices (these devices can represent a next generation of computers) and autonomous self-maintaining sensor networks. Such \index{sensor networks} sensor networks are closely related to autonomous underwater systems and cooperative micro- unmanned aerial vehicles. This application domain is very attractive for collective (especially swarms) systems, due to capabilities of converging large areas, fast reaction in case of accidence in monitored areas, lower power consumption, and self-deployment. Search and rescue scenarios can be also included here.

\textbf{Medical and biological applications.} \index{Medical and Health Care robotics} Medical and Health Care robotics as well as robo- companions for elderly people is a growing technological area. In many aspects, such systems overlap with service applications, however due to specific requirements, they are driven by different social and political forces. Biological applications are closely related to the medial domain, for example exploiting living bacteria for such activities as drug delivery or micro-actuation inside of human body. More generally, there are two approaches, which are actively developing here: making robots smaller, and making robot actuation more accurate and precise for nano-manipulation, whereas the robot itself remains large. This is a large and promising area, which has many potential applications.

\textbf{Service applications.} This field represents flexible robotic systems in warehouses, autonomous cleaning in large areas, diverse tasks in hospitals, guidance in museums, and various domestic and "@home" tasks~\cite{Iossifidis2004a}. To some extent, these applications are one of direct spin-offs from the \index{RoboCup} RoboCup initiative. Since these systems operate in unstructured environments and involve communication and interactions with human, a large amount of research is still required for successful applications. Since one of the main requirements in the service area is the adaptivity and flexibility of robots, there is a large interest for collective reconfigurable and self-developing robotics. In the following section we introduce this field in more detail.

\subsection{Service Area and Robot Companions}

Service area is directly related to interactions with humans and to the idea of \index{robot companions} robot companions. Overviewing the literature human-robot interactions, e.g.~\cite{Dautenhahn05}, \cite{WilkesAPRPK98}, \cite{Oestreicher06}, we can make several conclusions towards expectations of humans about a robot assistance. For instance, "In general it turned out that people were positive towards having robot assistance in their homes, ... between 30 and 50 \% of the informants said
that they were positive to have a robot support"~\cite{Oestreicher06}. Moreover in this work, several household tasks are identified such as dish washing (43\%), window polishing (39\%), dusting and other similar cleaning tasks (44\%), wet cleaning (of floors, etc.), washing clothes (37\%). Thus, the expectation is primarily related to not fully autonomous robots from the service-class platforms. In another work~\cite{Khan1998}, "The tasks that received most "NO"s [for robot assistance] were in order of frequency: baby sitting, reading aloud, watching cat/dog, acting as a butler, cooking food and taking care of kitchen goods. The tasks that received the most "ROBOT" meaning that a person actually wants a robot to help or conduct these tasks were in order of frequency: polishing windows, cleaning ceilings and walls, cleaning, wet cleaning, moving heavy things and wiping surfaces clean." Related to the appearance of robot platform (i.e. the class of robot platforms) the responses were 22\% neutral, 57\% machine-like and 19\% human-like design, however 71\% of subjects in~\cite{Dautenhahn05} prefer a communication with robot in human-like manner. As we can see, the relevance of humanoid platforms for service tasks is limited.

\begin{figure}[htp]
\centering
\subfigure[\label{fig:social1}]{\epsfig{file=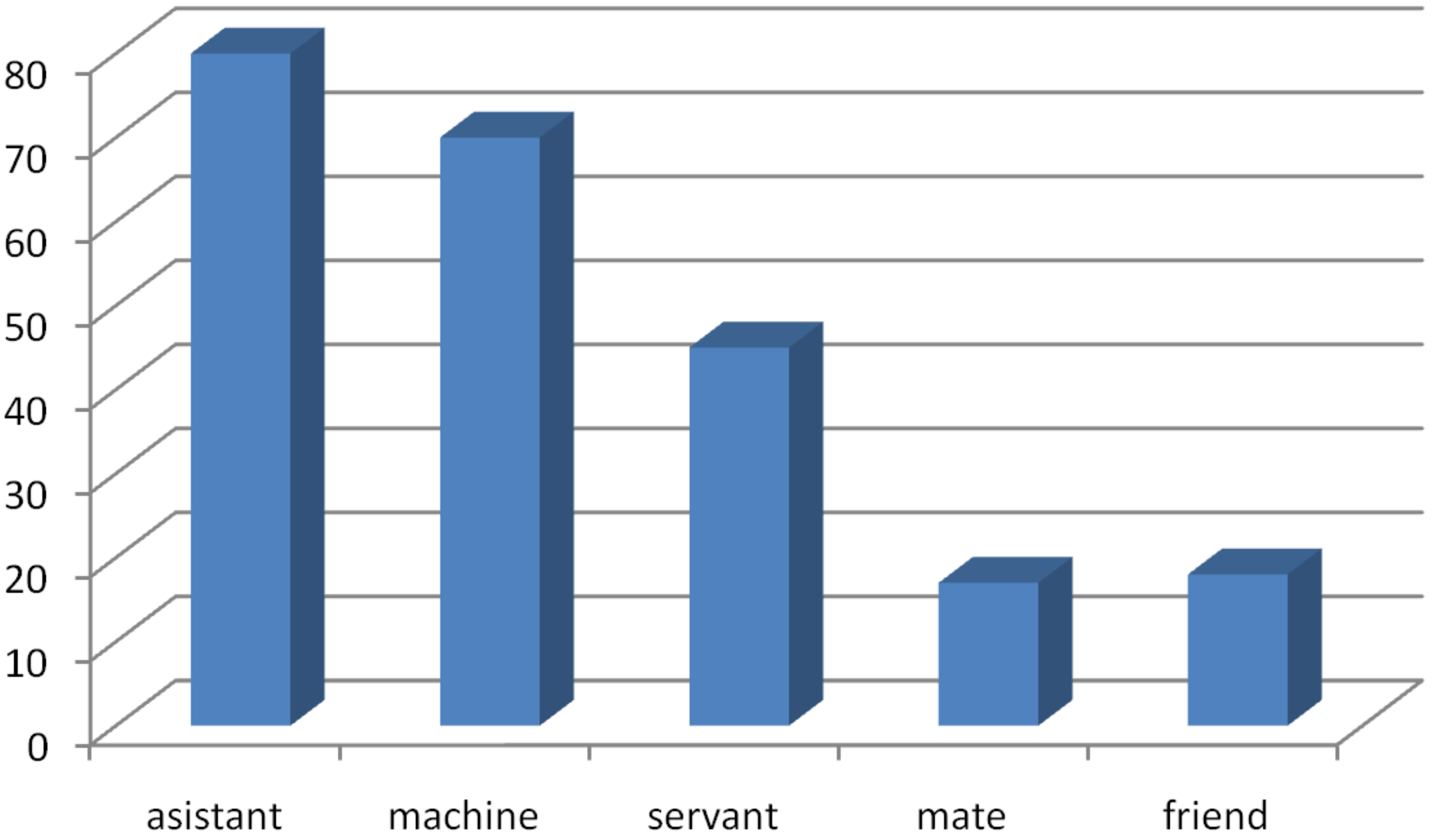,width=.49\textwidth}}~
\subfigure[\label{fig:social2}]{\epsfig{file=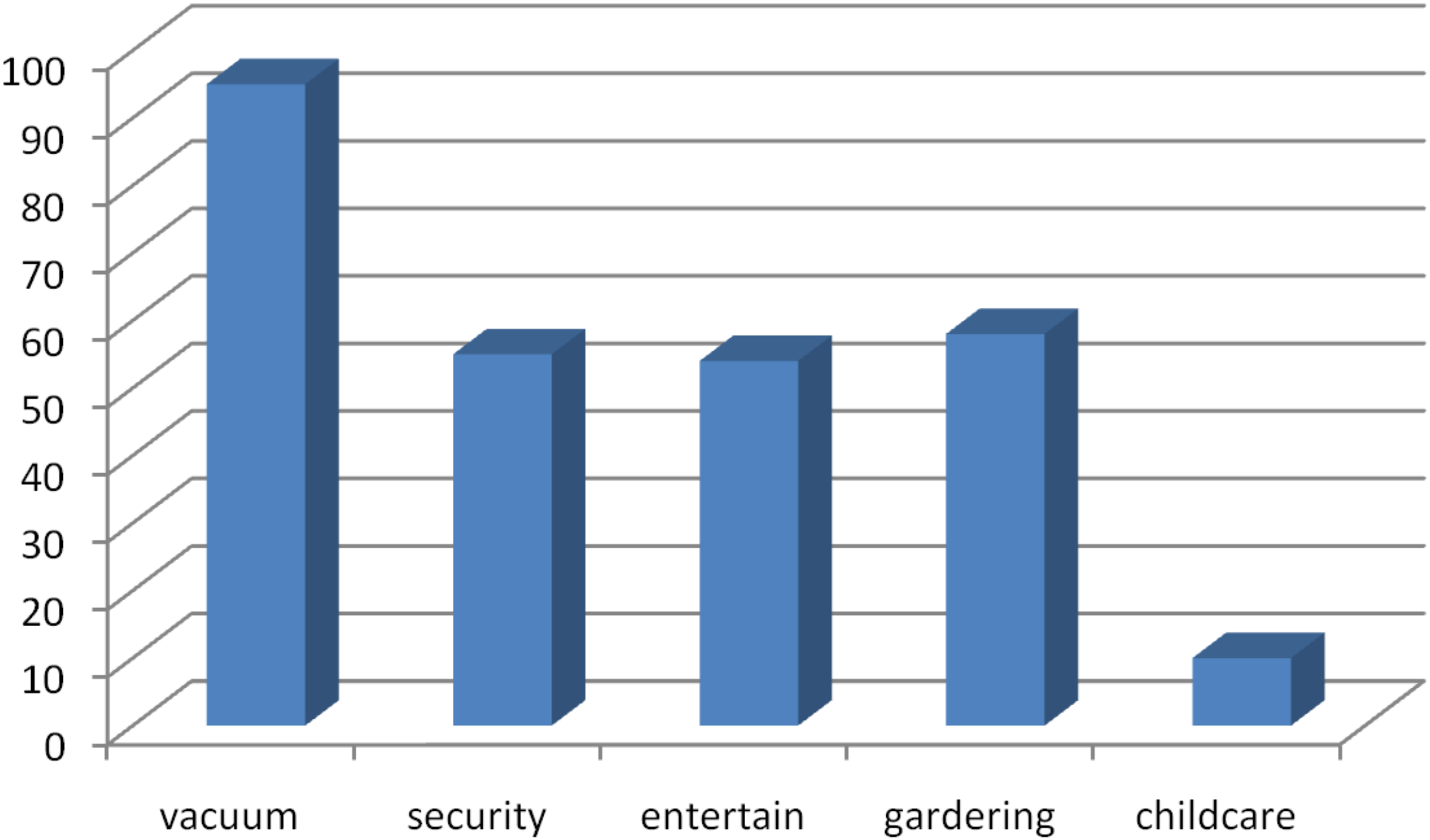,width=.49\textwidth}}
\caption{\textbf{(a)} Desired roles for a future robot companion and \textbf{(b)} preferred tasks for robot companion in the home (data from \cite{Dautenhahn05}).
\label{fig:social}}
\end{figure}

In the work~\cite{Dautenhahn05}, authors receive similar data, for example about 40\% of subjects liked having a robot companion at home. We reproduce two images related to preferred tasks for robot companion in the home and desired roles for a future robot companion from this work in Fig.~\ref{fig:social}.
Authors here came mainly to the conclusion that "The finding that people frequently cited that they would like a future robot to perform the role of a servant is maybe similar to the human 'butler' role". As we can see, understanding main facets of "@home applications" and the role of robot assistances is not essentially changed during last ten years. To some extent it also reflects a human desire for servants, which we can observe in all cultures and epochs.

\section{Conclusion}
\label{sec:conclusion}

The previous sections of this chapter are devoted to the different spectacular areas of modern robotics. We shortly overviewed key soft- and hardware technologies as well as considered fields of mechatronic, hybrid and collective robotics. We introduced robotics as a tool to assist humans and then made a revision of modern human's expectations towards such assistance. It turned out that robotics is driven by four main forces: industry, networked computing, service and bio-medical applications, whereas the main human's expectation remains to have a robot as a servant.

Unfortunately, many relevant points of robotics remained outside of this consideration. For example, cognitive aspects: sensing by different robots and problems of sensor-fusion, world modeling and collective localization. Another issue is learning: team- and group-based learning, collective and learning, and so on. We did not consider such points as mechanical and electrical engineering and design, which are very relevant for embodiment and, finally, for a success in applications. Many different research projects are addressing these issues, and they, due to complexity and amount of collected results, cannot be represented in one overview.

Considering these topics, we can make an interesting conclusion. When originally robotics was driven by humans, their needs and expectations, today in many branches of robotics, the technology is driven by its own motivations. It is related to improvement of sensors, actuators, scalability, redundancy or costs factors, whereas principal questions of usefulness in applications remained frequently outside of consideration. The technology is becoming "self-driven". To some extent this is caused by a human desire to create "life-like" systems in order to understand the life itself. Examples can be given by cognitive robotics and its close link to neuroscience and psychology.

These "self-driven" trends in robotics lead to so-called \index{self-* features} "self-issues" - self-adaptation, self-repairing, self-replication, self-reflection and similar. These self-issues are related in many aspects to adaptability and evolve-ability, to emergence of behavior and to controllability of long-term developmental processes. Currently, only a low effort is invested in understanding predictability of self-processes, principles of making purposeful self-developmental systems and consequences of a long-term independency and autonomy. We are facing even now an appearance of independent self-development of robots, e.g. in the area of evolutionary robotics; in future these processes will be much larger and intensive. We should be ready for emergence of \index{artificial robot cultures } artificial robot cultures and start to realize their meaning and consequences for the human.

\small

\end{document}